\documentclass[runningheads]{llncs}

\usepackage[mobile]{eccv}

\usepackage{eccvabbrv}

\usepackage{graphicx}
\usepackage{booktabs}
\usepackage{colortbl}
\usepackage{multirow}
\usepackage{makecell}

\usepackage[accsupp]{axessibility}  % Improves PDF readability for those with disabilities.

\usepackage[pagebackref=false,breaklinks=true,colorlinks,citecolor=eccvblue,bookmarks=true,bookmarksnumbered=true]{hyperref}

\usepackage{hyperref}

\usepackage{orcidlink}
\usepackage{marvosym}
\usepackage[absolute]{textpos}

\newcommand{\upperf}[1]{$^{\textcolor{red}{(\uparrow #1)}}$}
\definecolor{mypink}{rgb}{.99,.91,.95}

\begin{document}

% \title{Enhancing Vectorized HD Map Perception with Historical Rasterized Map} 
\title{Leveraging Enhanced Queries of Point Sets for Vectorized Map Construction}

\definecolor{somegray}{gray}{0.5}
\newcommand{\darkgrayed}[1]{\textcolor{somegray}{#1}}
\begin{textblock}{11}(2.5, -0.1)  %
\begin{center}
\darkgrayed{This paper has been accepted for publication at the European Conference on Computer Vision (ECCV), 2024.}
\end{center}
\end{textblock}

\titlerunning{MapQR}

\author{
Zihao Liu\inst{1, *}\orcidlink{0009-0004-6680-5319} \and
Xiaoyu Zhang\inst{2, *}\orcidlink{0000-0002-8674-3790} \and
Guangwei Liu\inst{3, *}\orcidlink{0009-0004-1585-8090} \and \\
Ji Zhao\inst{3}\textsuperscript{,\Letter}\orcidlink{0000-0002-0150-4601} \and
Ningyi Xu\inst{1}\textsuperscript{,\Letter}\orcidlink{0009-0004-6809-7694}
}

\authorrunning{Z.~Liu et al.}

\institute{Shanghai Jiao Tong University \and
The Chinese University of Hong Kong \and
Huixi Technology\\
\email{zhaoji84@gmail.com} \quad
\email{xuningyi@sjtu.edu.cn}
}

\maketitle

\renewcommand{\thefootnote}{}
\footnotetext{*: Equal contribution. \Letter: Corresponding author.}

\newcommand{\name}{MapQR}

\begin{abstract}
In autonomous driving, the high-definition (HD) map plays a crucial role in localization and planning.
Recently, several methods have facilitated end-to-end online map construction in DETR-like frameworks. However, little attention has been paid to the potential capabilities of exploring the query mechanism for map elements.
This paper introduces \emph{MapQR}, an end-to-end method with an emphasis on enhancing query capabilities for constructing online vectorized maps. 
To probe desirable information efficiently, MapQR utilizes a novel query design, called \emph{scatter-and-gather} query, which is modelled by separate content and position parts explicitly.
The base map instance queries are scattered to different reference points and added with positional embeddings to probe information from BEV features.
Then these scatted queries are gathered back to enhance information within each map instance.
Together with a simple and effective improvement of a BEV encoder, the proposed MapQR achieves the best mean average precision (mAP) and maintains good efficiency on both nuScenes and Argoverse~2.
In addition, integrating our query design into other models can boost their performance significantly.
The source code is available at \url{https://github.com/HXMap/MapQR}.
\end{abstract}
    
\section{Introduction}
\label{sec:intro}

High-definition (HD) maps, essential for autonomous driving, encapsulate precise vectorized details of map elements such as pedestrian crossing, lane divider, road boundaries, \textit{etc}.
As a fundamental element of autonomous systems, these systems capture imperative road topology and traffic rules to support the navigation and planning of vehicles.
Traditional SLAM-based HD map construction~\cite{zhang2014loam,shan2018lego} presents challenges such as complex pipelines, high costs, and notable localization errors.
Manual annotation further intensifies the labor and time demands. 
These limitations are prompting a shift towards online, learning-based methods that utilize onboard sensors.

Many existing works~\cite{li2022hdmapnet, zhou2022cross} define map construction as a semantic segmentation task in the bird's eye view (BEV) space, producing rasterized maps. 
They face limitations due to extensive post-processing required to acquire vectorized information.
To overcome the limitations of segmentation-based methods, new approaches have arisen that predict point sets to construct maps, utilizing a DETR-like structure for end-to-end map construction~\cite{liu2023vectormapnet, liao2022maptr, liao2023maptrv2, qiao2023end, ding2023pivotnet,yu2023scalablemap,liu2024mgmap,zhou2024himap}.

\begin{figure}[t] 
\centering 
\includegraphics[width=\linewidth]{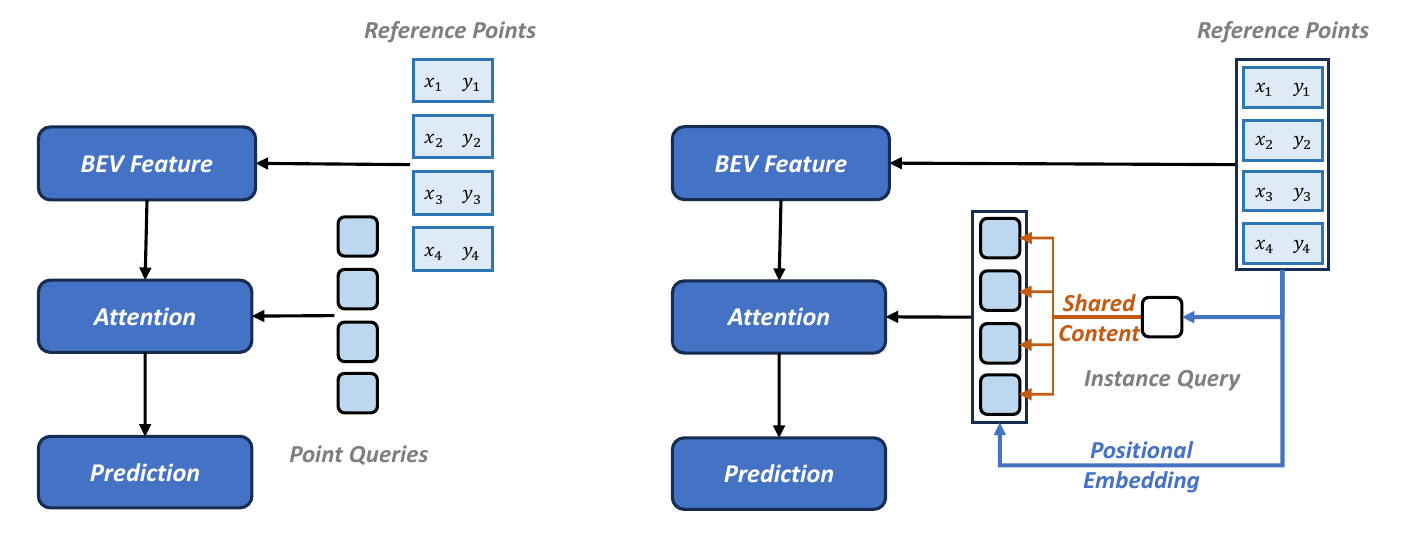}
\caption{Comparison of overall architectures. 
{\bf Left}: A DETR-like architecture exploited in many map construction methods. 
{\bf Right}: The proposed architecture with \emph{scatter-and-gather} query and \emph{positional embedding}.
Each instance is explicitly modelled by shared content part and different position parts.
The information within a single instance is also enhanced by the gathering operation.
In addition, reference points are used for positional embedding of these queries.}
\label{fig:teaser_1}
\end{figure}

The detection transformer (DETR)~\cite{carion2020end} is a transformer-based object detection architecture~\cite{vaswani2017attention}, in which learnable object queries are used to probe desirable information from image features. While the role of these learnable queries is still being studied, a common consensus is that a query consists of a semantic content part and a positional part. Thus corresponding objects can be recognized and localized in images~\cite{meng2021conditional, liu2022dab, li2022dn, zhang2022dino}. 
In Conditional DETR~\cite{meng2021conditional} and DAB-DETR~\cite{liu2022dab}, the positional part is explicitly encoded from reference points or box coordinates, which would then not be coupled with the content part and facilitate the learning process for both parts separately. These papers~\cite{meng2021conditional,liu2022dab} inspired us to design a proper query for point set prediction in the online map construction task.

DETR-like object detection methods predict a 4-dimensional bounding box information for each object by a set of learnable queries. 
As proved in DAB-DETR~\cite{liu2022dab}, each query in a decoder is composed of a decoder embedding (content information) and a learnable query (positional information).
By contrast, HD map construction usually predicts a point set for each map instance. 
In many state-of-the-art (SOTA) methods \cite{liao2022maptr,liao2023maptrv2, ding2023pivotnet}, point queries are used to probe information from BEV features, and each one predicts a point position. 
These predicted points are finally grouped to form the detected map elements.
As a result, the relation among point queries for the same element lacks explicit modelling, which would impede efficient learning for desirable content information.
In addition, current SOTA methods ignore the explicit location information and simply use learnable queries that are randomly initialized.

To solve the aforementioned problems, we propose to use instance queries rather than point queries and add positional embedding, as shown in \cref{fig:teaser_1}. 
Instead of predicting one position from each point query separately, we predict $n$ point positions simultaneously from each instance query.
To probe information from specific positions of BEV features, we generate positional embeddings from reference points like in Conditional DETR~\cite{meng2021conditional} (i.e., $n$ positional embeddings for each instance query). 
Then $n$ different positional embeddings will be added to one instance query, which becomes $n$ scattered queries for each instance.
Therefore, each map element contains a group of scattered queries that share the same content part from a single instance query and different positional parts embedded from reference positions. 
The set of scattered queries is gathered back as an instance query to enhance information within each instance.
We call this query a \emph{scatter-and-gather} query.
The relations within a single instance is explicitly modelled as shared content and different positions, which are enhanced by scattering and gathering.
It also allows the proposed decoder to increase the number of queries to achieve higher accuracy without noticeably increasing the computational burden and memory usage.
The query design is the fundamental cornerstone of our proposed method and, with a simple and effective improvement of a BEV encoder, constitutes our proposed \emph{MapQR}.

We have performed extensive experiments to demonstrate the superiority of the proposed method MapQR.
Our method has superior performance in both nuScenes and Argoverse 2 map construction tasks while maintaining good efficiency.
Furthermore, we integrated the fundamental MapQR design into other SOTA models, substantially improving their final performance.
In summary, the contributions of this paper are three-fold.
\begin{itemize}
	\item We proposed an online end-to-end map construction method which is based on a novel \emph{scatter-and-gather} query. This query design, coupled with a compatible positional embedding, is beneficial for point-set-based instance detection within DETR-like architectures. 
	\item The proposed online map construction method performs better on existing online map construction benchmarks than prior arts.
	\item The incorporation of our core design into other SOTA online map construction methods also yields remarkable improvements in accuracy.
\end{itemize}

\section{Related Work}
\label{sec:relwork}

\subsection{Online Vectorized HD Map Construction}
HD map is designed to encapsulate precise vectorized representation of map elements for autonomous driving. While conventional offline SLAM-based methods~\cite{zhang2014loam, shan2018lego} suffer from high maintaining cost, online HD map construction draws increasing attention. Online map construction can be solved by segmentation~\cite{philion2020lift, li2022bevformer, zhou2022cross, Gosala_2023_CVPR} or lane detection~\cite{Feng_2022_CVPR, chen2022persformer, Wang_2023_CVPR} in BEV space, producing rasterized maps. HDMapNet~\cite{li2022hdmapnet} makes a further step of grouping segmentation results to vectorized representations. 

VectorMapNet~\cite{liu2023vectormapnet} proposes the first end-to-end vectorized map learning framework and utilizes auto-regression to predict points sequentially. Later end-to-end methods for vectorized HD map construction become popular \cite{liao2022maptr,liao2023maptrv2,shin2023instagram,qiao2023end,ding2023pivotnet,xu2023insightmapper,yu2023scalablemap,yuan2024streammapnet_wacv,zhang2023online,liu2024mgmap,zhou2024himap}. MapTR~\cite{liao2022maptr} treats online map construction as a point set prediction problem and designs a DETR-like framework~\cite{carion2020end}, achieving state-of-the-art performance. Its improved version MapTRv2~\cite{liao2023maptrv2} designs a decoupled self-attention decoder and auxiliary losses to improve the performance further. Instead of representing a map element by a point set, BeMapNet~\cite{qiao2023end} uses piecewise B{\'e}zier curve, and PivotNet~\cite{ding2023pivotnet} uses pivot-based points for more elaborate modeling. 
Most of these works use point-based queries and should meet the problem stated in \cref{sec:intro}. 
Recently, StreamMapNet~\cite{yuan2024streammapnet_wacv} utilizes multi-point attention for a wide perception range and exploits long-sequence temporal fusion. Unlike previous work, we provide an in-depth exploitation of the instance-based query and design scatter-and-gather query to probe both content and positional information accurately.

\subsection{Multi-View Camera-to-BEV Transformation}
HD map is commonly constructed under BEV view, and thus online HD map construction methods depend on converting visual features from camera perspective to BEV space. LSS~\cite{philion2020lift} utilizes latent depth distribution to lift 2D image features to 3D space, and uses pooling later to aggregate to BEV features. BEVformer~\cite{li2022bevformer} depends on the transformer architecture~\cite{vaswani2017attention} and designs spatial cross-attention based on visual projection. Unlike deformable attention~\cite{dai2017deformable} used in BEVformer, GKT~\cite{chen2022efficient} designs a geometry-guided kernel transformer. 
All these methods can produce BEV features effectively, and our proposed decoder is compatible with all of them.

\subsection{Detection Transformers}

DETR~\cite{carion2020end} builds the first end-to-end object detector with transformer~\cite{vaswani2017attention}, eliminating the need for many hand-designed components. 
In the object detection task, it represents the object boxes as a set of queries, and directly uses the transformer encoder-decoder architecture to interact the queries with the image to predict the set of bounding boxes.
Subsequent works \cite{zhu2021deformable, zhu2021deformable, meng2021conditional, wang2022anchor, liu2022dab, li2022dn, zhang2022dino} further optimize this paradigm and performance. In Deformable DETR~\cite{zhu2021deformable}, multi-scale features are incorporated to solve the problem of poor detection results of the vanilla DETR for small objects. In Conditional DETR~\cite{meng2021conditional} and DAB-DETR~\cite{liu2022dab}, the convergence of DETR is accelerated by adding positional embedding to learnable queries. 
In DN-DETR~\cite{li2022dn} and DINO~\cite{zhang2022dino}, the actual boxes with noise are input into the decoder, so as to establish a more stable pairing relationship to accelerate convergence.
The model in our method is a DETR-like model. Leveraging point set representation of map elements, we design queries that can scatter and gather in a transformer decoder, and integrate positional information into the initialization of queries.

\section{Method}
\label{sec:method}

\subsection{Overall Architecture}
\begin{figure*}[tbp] 
\centering 
\includegraphics[width=0.95\linewidth]{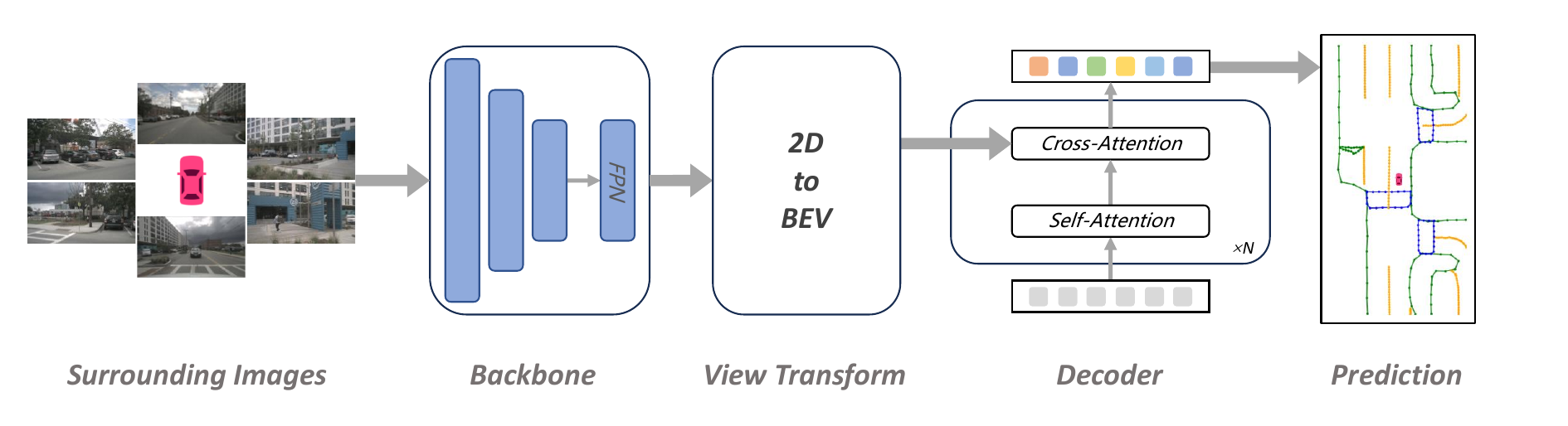}
\caption{The overall architecture of our method. It contains three main components: a shared image backbone to extract image features, a view transformation module to obtain BEV features, and a transformer decoder for generating predictions.
The backbone and view transformation modules can be any popular one without additional adaption.
The decoder is our key design, and in principle it can be directly applied to other DETR-like models of map construction.}
\label{fig:architecture}
\end{figure*}

The model of our method takes sequences of multi-view images as inputs to construct an HD map end-to-end, with the objective of generating sets of predicted points to represent instances of map elements.
Each instance of a map element comprises a class label and a set of predicted points. Each predicted point includes explicit positional information to create polylines that represent the shape and location of the instance.
The overall architecture of our model is shown in \cref{fig:architecture}, which is similar in structure to other end-to-end models~\cite{liu2023vectormapnet,liao2022maptr}.

%%%%%%%%%%%%%%%%%
%%%%%%%%%%%%%%%%%

Given surrounding images from a multi-camera rig as inputs, we first extract image features through a shared 2D backbone. 
These image features of the cameras’ perspective view (PV) are converted to BEV representations by feeding into a view transformation module, commonly referred to as BEV encoder. The resulted BEV features are denoted as $F_{\mathrm{bev}} \in \mathbb{R}^{C \times H \times W}$, where $C$, $H$, $W$ represent the number of feature channels, height, and width, respectively.
All mainstream 2D backbones~\cite{he2016deep, liu2021swin} and view transformation modules which can be used for PV-to-BEV transformation~\cite{chen2022efficient, li2022bevformer} can be integrated into our model.

\begin{figure*}[t] 
\centering 
\includegraphics[width=0.95\linewidth]{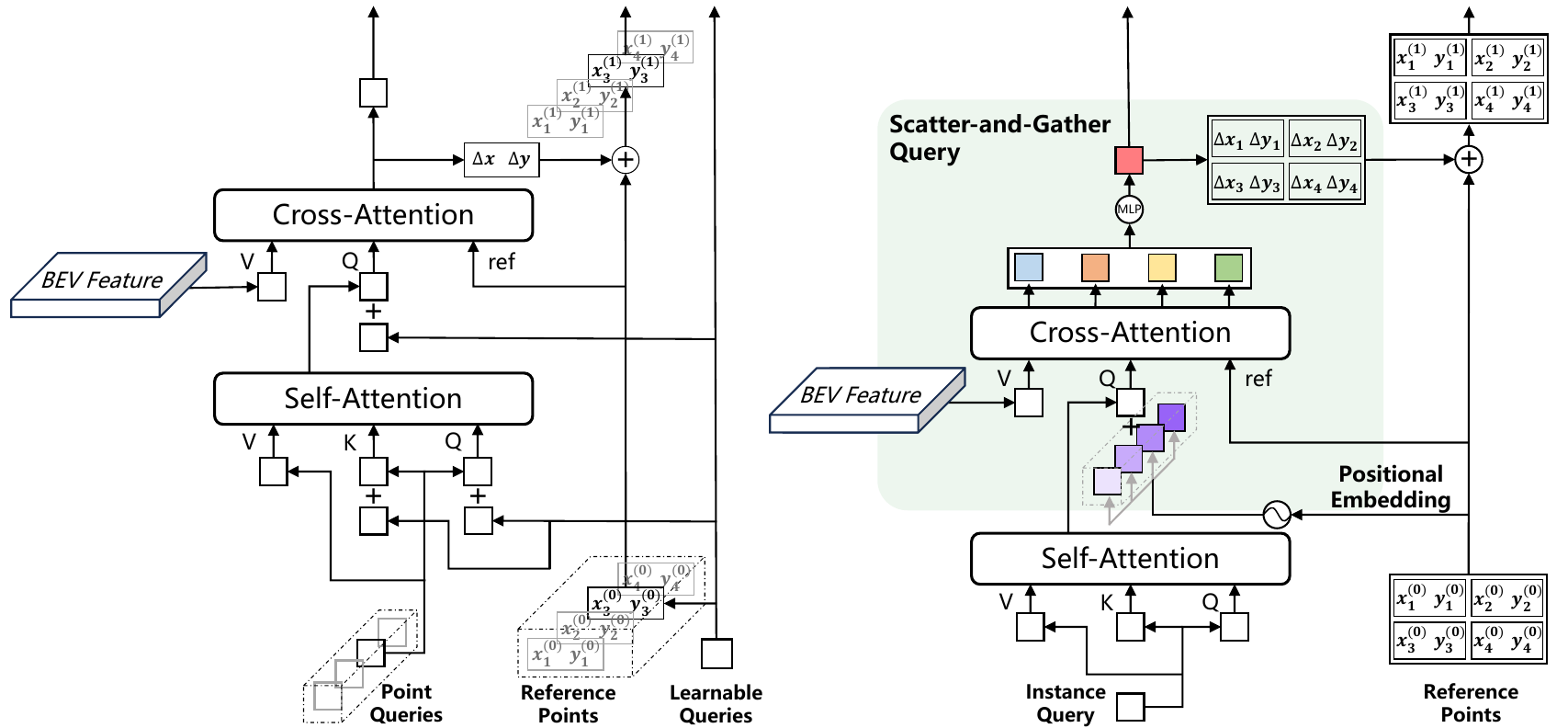}
\caption{Comparison of decoders. {\bf Left}: The decoder of MapTR~\cite{liao2022maptr}. {\bf Right}: The proposed decoder of MapQR. In this example, $4$ reference points are contained in an instance.
}
\label{fig:decoder_vs}
\end{figure*}

\subsection{Decoder with Scatter-and-Gather Query}
The decoder of the architecture is our core design, consisting of stacked transformer layers, with its detail shown in \cref{fig:decoder_vs}.
The improvements of our decoder mainly revolve around the query design, including the {\emph{scatter-and-gather}} query and its compatible positional embedding.

{\bf Scatter-and-Gather Query.}
In our method, we propose using only one query type, instance query, as input to the decoder.
We define a set of instance queries $\left\{q_i^{\mathrm{ins}}\right\}_{i=1}^N$, and each map element (with index $i$) corresponds to an instance query $q_i^{\mathrm{ins}}$.
$N$ represents the number of instance queries, configured to exceed the typical number of map elements presenting in a scene. 
The $i$-th instance query $q_i^{\mathrm{ins}}$ is copied into $n$ scattered queries $\left\{q_{i,j}^{\mathrm{sca}}\right\}_{j=1}^n$ by an operator $\operatorname{scatter}$, i.e., $\left\{q_{i,j}^{\mathrm{sca}}\right\}_{j=1}^n = \operatorname{scatter}(q_i^{\mathrm{ins}})$.
The operator $\operatorname{scatter}$ takes an input data vector and replicates it into multiple copies. Formally,
\begin{equation}
	\operatorname{scatter}(f)= \underbrace{\{f, f, \ldots, f\}}_{n \ \text{replicas}},
\end{equation}
where $n$ is the number of sequential points to model a map element.
After passing through a transformer layer, scattered queries $\left\{q_{i,j}^{\mathrm{sca}}\right\}_{j=1}^n$ are aggregated into a single instance query using a $\operatorname{gather}$ operator, which is composed of an MLP as following:
\begin{equation}
    \operatorname{gather}(\left\{f_j\right\}_{j=1}^n)=\operatorname{MLP}(\operatorname{concat}(\left\{f_j\right\}_{j=1}^n)).
\end{equation}
The operator $\operatorname{concat}$ represents concatenation.

A complete decoder layer consists of a self-attention (SA) module and a cross-attention (CA) module.
In each decoder layer, self-attention is used to exchange information between instance queries $\left\{q_i^{\mathrm{ins}}\right\}_{i=1}^N$.
The output of self-attention is scattered into multiple copies, namely scattered queries $\left\{q_{i,j}^{\mathrm{sca}}\right\}_{i=1,j=1}^{N,n}$, and is used as input to cross-attention.
The attention score is calculated with BEV features to probe the information from BEV space.

The detailed architectures of the proposed decoder with comparison to MapTR series~\cite{liao2022maptr,liao2023maptrv2} are demonstrated in \cref{fig:decoder_vs}. 
MapTR~\cite{liao2022maptr} uses point queries as the input of decoder layers and calculates self-attention scores between them.
This makes the computational complexity of self-attention much higher than our model because of much more involved queries. 
By contrast, the proposed query design can enhance information in the same map element.
Meanwhile, this design reduces the number of queries to reduce computational consumption in self-attention. Thus, it allows the involvement of more instance queries without significantly increasing memory consumption.
Detailed experimental results are reported in \cref{tab:query_num}.

{\bf Positional Embedding.}
As shown in \cref{fig:decoder_vs}, another key difference between the decoder of our model and other SOTA models is the usage of positional embedding.
As demonstrated in DAB-DETR~\cite{liu2022dab}, a query is composed of a content part and a positional part. 
By designing a scatter-and-gather scheme of instance query, scattered queries corresponding to the same map element can share content information. 
Then another aspect that deserves exploration should be the positional information.
In Conditional DETR~\cite{meng2021conditional} and DAB-DETR~\cite{liu2022dab}, the positional part is explicitly encoded from reference points or box coordinates so that it is not coupled to the content part and facilitates a learning process for both parts separately.

Considering the point set representation for the map construction task, we propose a more suitable way of positional embedding as shown in \cref{fig:decoder_vs}.
In a decoder layer, each instance query $q_i^\mathrm{ins}$ is scattered into $n$ scattered queries $\left\{q_{i,j}^{\mathrm{sca}}\right\}_{j=1}^n$ after passing through the self-attention module.
Each scattered query adds different positional embedding $P_{i,j} \in \mathbb{R}^D$ from its reference point $A_{i,j} = (x_{i,j}, y_{i,j})$, where $x_{i,j}, y_{i,j} \in \mathbb{R}$ are position coordinates, and $D$ is the dimension of query and positional embedding.
The positional embedding $P_{i,j}$ of $j$-th scattered query is generated by:
\begin{equation}
P_{i,j}=\operatorname{LP}\left(\operatorname{PE}\left(A_{i,j}\right)\right),
\label{eq:pe}
\end{equation}
where $\operatorname{LP}$ is a learnable linear projection, and $\operatorname{PE}$ means positional encoding to generate sinusoidal embeddings from float numbers.
We overload the $\operatorname{PE}$ operator in the same way as DAB-DETR~\cite{liu2022dab} to compute positional encoding of $A_{i,j}$:
\begin{equation}
\operatorname{PE}\left(A_{i,j}\right)=\operatorname{concat}\left(\operatorname{PE}\left(x_{i,j}\right), \operatorname{PE}\left(y_{i,j}\right)\right).
\end{equation}
The positional encoding function $\operatorname{PE}$ maps a float number into a vector of $D/2$ dimensions as $\mathbb{R} \rightarrow \mathbb{R}^{D / 2}$, and thus $\operatorname{PE}\left(A_{i,j}\right) \in \mathbb{R}^D$.

We have also tried adding positional embedding to instance query $q_i^\mathrm{ins}$ before self-attention modules, such as using learnable queries or encoding from the smallest bounding box covering the set of reference points. However, it did not achieve positive results, which are included in the supplementary material. So, the positional embedding of the instance query is excluded in our model.

In summary, the calculation for $i$-th instance query in the scatter-and-collection query mechanism is
\begin{equation}
	\label{equ:f_decoder}
 \operatorname{gather}\left(\operatorname{CA}\left(\operatorname{scatter}\left(\operatorname{SA}\left(q_i^{\mathrm {ins }}\right)\right) + \left\{P_{i,j}\right\}_{j=1}^{n}, F_\mathrm{bev}\right)\right).
\end{equation}
where $\left\{P_{i,j}\right\}_{j=1}^{n}$ are the set of positional embeddings. The resulted instance queries will be matched with the map elements to calculate the loss.

\subsection{BEV Encoder: GKT with Flexible Height}
\label{sec:gkt-star}

BEV encoders learn to adapt to 3D space implicitly or explicitly. For example, BEVFormer~\cite{li2022bevformer} samples 3D reference points from fixed heights for each BEV query while learning adaptable offsets from projected 2D positions. Similar to BEVFormer, GKT \cite{chen2022efficient} uses a fixed kernel to extract features around projected 2D image positions. Since GKT depends on a fixed 3D transformation, it lacks certain flexibility. Still, it has remarkable performance compared with BEVFormer~\cite{liao2022maptr}. 

To further increase the flexibility of GKT, we predict adaptive height offsets to original fixed heights when generating 3D reference points $\boldsymbol{p}_i = (x_i^\mathrm{bev}, y_i^\mathrm{bev}, z_i^\mathrm{bev})$ for each BEV query $q_{i}^{\mathrm{bev}}$:
\begin{equation}
z_i^\mathrm{bev} = z_0^\mathrm{bev} + \operatorname{LP}(q_{i}^{\mathrm{bev}}),
\end{equation}
where offsets are predicted from query $q_{i}^{\mathrm{bev}}$ using a learable linear projection $\operatorname{LP}$, and 3D points $(x_i^\mathrm{bev}, y_i^\mathrm{bev}, z_0^\mathrm{bev})$ are sampled from BEV space as BEVFormer~\cite{li2022bevformer}.
After projecting to 2D images, we still use a fixed kernel to extract features. We find this simple modification can further improve prediction results. It even outperforms BEVPoolv2~\cite{huang2022bevpoolv2}, which is used with auxiliary depth loss in MapTRv2. 
Its superiority with ablation study is demonstrated in experiments. We denote this improved GKT as {\bf GKT-h}, in which the suffix `h' stands for height.

\subsection{Matching Cost and Training Loss}
Our method uses hierarchical bipartite matching similar to the MapTR series~\cite{liao2022maptr,liao2023maptrv2}.
It is necessary to find the optimal instance assignment $\hat{\pi}$ between predicted map elements $\left\{\hat{y}_i\right\}_{i=1}^N$ and ground truth (GT) map elements $\left\{{y}_i\right\}_{i=1}^N$ as follows:
\begin{equation}
\hat{\pi}=\underset{\pi \in \Pi_N}{\arg \min } \sum_{i=1}^N \mathcal{L}_{\mathrm{ins\_match }}\left(\hat{y}_{\pi(i)}, y_i\right),
\end{equation}
where $\Pi_N$ is the set of permutations with $N$ elements.
$\mathcal{L}_{\mathrm{ins\_match }}\left(\hat{y}_{\pi(i)}, y_i\right)$ is a pair-wise matching cost between predictions $\left\{\hat{y}_i\right\}$ and GT map elements $\left\{{y}_i\right\}$.
Then we use matching of scattered instances to find an optimal point-to-point assignment between the set of scattered instances and the set of GT points. 
The matching is established by calculating the Manhattan distance.

The loss function is mainly consistent with MapTRv2~\cite{liao2023maptrv2}. 
It is composed of three parts:
\begin{equation}
\mathcal{L}=\beta_o \mathcal{L}_{\text {one2one}}+\beta_m \mathcal{L}_{\text {one2many}}+\beta_d \mathcal{L}_{\text {dense}},
\end{equation}
where $\beta_o$, $\beta_m$, and $\beta_d$ are the weights for balancing different loss terms. We set the same values as MapTRv2.
The basic loss function $\mathcal{L}_{\text {one2one}}$ consists of three components: classification loss, point-to-point loss, and edge direction loss.
Similarly, we repeat the ground-truth map elements $K$ times and fill them with the empty set $\varnothing$ to form a set of size $T$ to compute the auxiliary one-to-many set prediction loss $\mathcal{L}_{\text {one2many}}$.
The auxiliary dense prediction loss $\mathcal{L}_{\text {dense}}$ differs from MapTRv2 due to its composition of two parts, as the BEV encoder of our method no longer predicts depth. Formally,
\begin{equation}
\mathcal{L}_{\text {dense }}=\alpha_b \mathcal{L}_{\text {BEVSeg}}+\alpha_p \mathcal{L}_{\text {PVSeg}},
\end{equation}
where definitions of BEV segmentation loss and PV segmentation loss are the same as MapTRv2.

Since the proposed method mainly uses a novel design of query, we call the proposed method (scatter-and-gather query + positional embedding + GKT-h) as {\bf MapQR}. Here QR is short for query. 
Sometimes we also want to emphasize the core contribution, i.e., the scatter-and-gather query with positional embedding. This part is denoted as {\bf SGQ} for short.

\section{Experiments}
\label{sec:exp}

We compare the proposed method {MapQR} with SOTA methods, including MapTR series~\cite{liao2022maptr,liao2023maptrv2}, BeMapNet~\cite{qiao2023end},  PivotNet~\cite{ding2023pivotnet} and StreamMapNet~\cite{yuan2024streammapnet_wacv}.
To validate the effectiveness of the proposed {SGQ} decoder separately, 
we also integrate it into the MapTR series and StreamMapNet by replacing their point-query-based decoders with the proposed decoder. 

\subsection{Settings}
{\bf Datasets.} We evaluate our method on public datasets nuScenes~\cite{caesar2020nuscenes} and Argoverse~2 \cite{wilson2021argoverse}. These datasets provide surrounding images captured by self-driving cars in hundreds of city scenes. Each sample contains $6$ RGB images in nuScenes and $7$ RGB images in Argoverse~2. nuScenes provides 2D vectorized map elements, and Argoverse 2 provides 3D vectorized map elements as ground truth.

{\bf Evaluation Metric.} Following previous works~\cite{li2022hdmapnet, liao2022maptr}, three static map categories are used for performance evaluation, i.e.,  lane divider, pedestrian crossing, and road boundary. Chamfer distance is used to determine whether the prediction matches the ground truth. Besides the thresholds $\{0.5m, 1.0m, 1.5m\}$ used in MapTR~\cite{liao2022maptr}, the results are also evaluated with smaller thresholds $\{0.2m, 0.5m, 1.0m\}$ used in BeMapNet~\cite{qiao2023end}. The corresponding mAP metrics are denoted as mAP$_2$ and mAP$_1$, respectively.

%%%%%%%%%%%%%%%%%
\definecolor{mygray}{rgb}{0.9,0.9,0.9}

\begin{table}[t]
  \centering
  \caption{Comparison with SOTA methods on nuScenes with smaller thresholds \{0.2m, 0.5m, 1.0m\}. FPS is tested on the same NVIDIA 4090 machine. The results in \colorbox{mygray}{gray} are taken from corresponding papers directly, and others are reproduced by ourselves. The best results for the same settings (i.e., backbone and epoch) are labeled in bold.}
    \begin{tabular}{c|cc|cccc|c}
    \toprule
    \makebox[3cm][c]{Method} & \makebox[1.3cm][c]{Backbone} & \makebox[1cm][c]{Epoch} & \makebox[1.2cm][c]{AP$_{div}$} & \makebox[1.2cm][c]{AP$_{ped}$} & \makebox[1.2cm][c]{AP$_{bou}$} & \makebox[1.2cm][c]{mAP$_1$}  & \makebox[1cm][c]{FPS} \\
    \midrule
    \multirow{3}{*}{BeMapNet \cite{qiao2023end}} & R50   & 30    & \cellcolor[rgb]{ .906,  .902,  .902}46.9 & \cellcolor[rgb]{ .906,  .902,  .902}39.0 & \cellcolor[rgb]{ .906,  .902,  .902}37.8 & \cellcolor[rgb]{ .906,  .902,  .902}41.3 & 8.3 \\
          & R50   & 110   & \cellcolor[rgb]{ .906,  .902,  .902}52.7 & \cellcolor[rgb]{ .906,  .902,  .902}44.5 & \cellcolor[rgb]{ .906,  .902,  .902}44.2 & \cellcolor[rgb]{ .906,  .902,  .902}47.1 & 8.3 \\
          & SwinT & 30    & \cellcolor[rgb]{ .906,  .902,  .902}49.1 & \cellcolor[rgb]{ .906,  .902,  .902}42.2 & \cellcolor[rgb]{ .906,  .902,  .902}39.9 & \cellcolor[rgb]{ .906,  .902,  .902}43.7 & 7.9 \\
    \midrule
    \multirow{2}{*}{PivotNet \cite{ding2023pivotnet}} & R50   & 24    & \cellcolor[rgb]{ .906,  .902,  .902}41.4 & \cellcolor[rgb]{ .906,  .902,  .902}34.3 & \cellcolor[rgb]{ .906,  .902,  .902}39.8 & \cellcolor[rgb]{ .906,  .902,  .902}38.5 & - \\
          & SwinT & 24    & \cellcolor[rgb]{ .906,  .902,  .902}45.0 & \cellcolor[rgb]{ .906,  .902,  .902}36.2 & \cellcolor[rgb]{ .906,  .902,  .902}41.2 & \cellcolor[rgb]{ .906,  .902,  .902}40.8 & - \\
    \midrule
    \multirow{4}{*}{MapTR \cite{liao2022maptr}} & R50   & 24    & 30.7  & 23.2  & 28.2  & 27.3  & 24.2 \\
          & R50   & 110   & 40.5  & 31.4  & 35.5  & 35.8  & 24.2 \\
          & swinT & 24    & 30.4  & 24.5  & 29.9  & 28.9  & 18.2 \\
          & SwinB & 24    & 36.5  & 29.0    & 35.2  & 33.5  & 9.3 \\
    \midrule
    \multirow{2}{*}{MapTRv2 \cite{liao2023maptrv2}} & R50   & 24    & 40.0    & 35.4  & 36.3  & 37.2  & 19.6 \\
          & R50   & 110   & 49.0    & 43.6  & 43.7  & 45.4  & 19.6 \\
    \midrule
    StreamMapNet \cite{yuan2024streammapnet_wacv} & R50   & 24    & 42.9  & 32.3  & 33.2  & 36.2  & 19.9 \\
    \midrule
    \multicolumn{1}{c|}{\multirow{4}{*}{\textbf{MapQR(Ours)}}} & R50   & 24    & 49.9  & 38.6  & 41.5  & \textbf{43.3} & 17.9 \\
          & R50   & 110   & 57.3  & 46.2  & 48.1  & \textbf{50.5} & 17.9 \\
          & swinT & 24    & 49.3  & 38.5  & 41.1  & \textbf{43.0} & 14.8 \\
          & swinB & 24    & 52.3  & 41.4  & 43.2  & \textbf{45.6} & 8.3 \\
    \bottomrule
    \end{tabular}
 \label{tab:nuscenes}
\end{table}

\begin{table}[t]
  \centering
  \caption{Comparison with SOTA methods on nuScenes with thresholds \{0.5m, 1.0m, 1.5m\}.
  FPS is tested on the same NVIDIA 4090 machine. The results in \colorbox{mygray}{gray} are taken from corresponding papers directly, and others are reproduced by ourselves. The best results for the same settings (i.e., backbone and epoch) are labeled in bold.
  }
    \begin{tabular}{c|cc|cccc|c}
    \toprule
    \makebox[3cm][c]{Method} & \makebox[1.3cm][c]{Backbone} & \makebox[1cm][c]{Epoch} & \makebox[1.2cm][c]{AP$_{div}$} & \makebox[1.2cm][c]{AP$_{ped}$} & \makebox[1.2cm][c]{AP$_{bou}$} & \makebox[1.2cm][c]{mAP$_2$}  & \makebox[1cm][c]{FPS} \\
    \midrule
    \multirow{3}{*}{BeMapNet\cite{qiao2023end}} & R50   & 30    & \cellcolor[rgb]{ .906,  .902,  .902}62.3 & \cellcolor[rgb]{ .906,  .902,  .902}57.7 & \cellcolor[rgb]{ .906,  .902,  .902}59.4 & \cellcolor[rgb]{ .906,  .902,  .902}59.8 & 8.3 \\
          & R50   & 110   & \cellcolor[rgb]{ .906,  .902,  .902}66.7 & \cellcolor[rgb]{ .906,  .902,  .902}62.6 & \cellcolor[rgb]{ .906,  .902,  .902}65.1 & \cellcolor[rgb]{ .906,  .902,  .902}64.8 & 8.3 \\
          & SwinT & 30    & \cellcolor[rgb]{ .906,  .902,  .902}64.4 & \cellcolor[rgb]{ .906,  .902,  .902}61.2 & \cellcolor[rgb]{ .906,  .902,  .902}61.7 & \cellcolor[rgb]{ .906,  .902,  .902}62.4 & 7.9 \\
    \midrule
    \multirow{2}{*}{PivotNet \cite{ding2023pivotnet}} & R50   & 24    & \cellcolor[rgb]{ .906,  .902,  .902}56.5 & \cellcolor[rgb]{ .906,  .902,  .902}56.2 & \cellcolor[rgb]{ .906,  .902,  .902}60.1 & \cellcolor[rgb]{ .906,  .902,  .902}57.6 & - \\
          & SwinT & 24    & \cellcolor[rgb]{ .906,  .902,  .902}60.6 & \cellcolor[rgb]{ .906,  .902,  .902}59.2 & \cellcolor[rgb]{ .906,  .902,  .902}62.2 & \cellcolor[rgb]{ .906,  .902,  .902}60.6 & - \\
    \midrule
    \multirow{4}{*}{MapTR \cite{liao2022maptr}} & R50   & 24    & \cellcolor[rgb]{ .906,  .902,  .902}51.5 & \cellcolor[rgb]{ .906,  .902,  .902}46.3 & \cellcolor[rgb]{ .906,  .902,  .902}53.1 & \cellcolor[rgb]{ .906,  .902,  .902}50.3 & 24.2 \\
          & R50   & 110   & \cellcolor[rgb]{ .906,  .902,  .902}59.8 & \cellcolor[rgb]{ .906,  .902,  .902}56.2 & \cellcolor[rgb]{ .906,  .902,  .902}60.1 & \cellcolor[rgb]{ .906,  .902,  .902}58.7 & 24.2 \\
          & swinT & 24    & \cellcolor[rgb]{ .906,  .902,  .902}45.2 & \cellcolor[rgb]{ .906,  .902,  .902}52.7 & \cellcolor[rgb]{ .906,  .902,  .902}52.3 & \cellcolor[rgb]{ .906,  .902,  .902}50.1 & 18.2 \\
          & SwinB & 24    & \cellcolor[rgb]{ .906,  .902,  .902}58.7 & \cellcolor[rgb]{ .906,  .902,  .902}50.6 & \cellcolor[rgb]{ .906,  .902,  .902}58.4 & \cellcolor[rgb]{ .906,  .902,  .902}55.9 & 9.3 \\
    \midrule
    \multirow{2}{*}{MapTRv2 \cite{liao2023maptrv2}} & R50   & 24    & \cellcolor[rgb]{ .906,  .902,  .902}62.4 & \cellcolor[rgb]{ .906,  .902,  .902}59.8 & \cellcolor[rgb]{ .906,  .902,  .902}62.4 & \cellcolor[rgb]{ .906,  .902,  .902}61.5 & 19.6 \\
          & R50   & 110   & \cellcolor[rgb]{ .906,  .902,  .902}68.3 & \cellcolor[rgb]{ .906,  .902,  .902}68.1 & \cellcolor[rgb]{ .906,  .902,  .902}69.7 & \cellcolor[rgb]{ .906,  .902,  .902}68.7 & 19.6 \\
    \midrule
    StreamMapNet \cite{yuan2024streammapnet_wacv} & R50   & 24    & 64.1  & 58.2  & 59.4  & 60.6  & 19.9 \\
    \midrule
    \multicolumn{1}{c|}{\multirow{4}{*}{\textbf{MapQR(Ours)}}} & R50   & 24    & 68    & 63.4  & 67.7  & \textbf{66.4} & 17.9 \\
          & R50   & 110   & 74.4  & 70.1  & 73.2  & \textbf{72.6} & 17.9 \\
          & swinT & 24    & 68.1  & 63.1  & 67.1  & \textbf{66.1} & 14.8 \\
          & swinB & 24    & 72.1  & 67.5  & 70.5  & \textbf{70.1} & 8.3 \\
    \bottomrule
    \end{tabular}
   \label{tab:nuscenes_2}
\end{table}

%%%%%%%%%%%%%%%%%

\begin{table}[h]
  \centering
  \caption{Comparison with SOTA methods on Argoverse 2. Map dim denotes the dimension used to model map elements. $dim=2$ indicates dropping the height of each predicted map elements, while $dim=3$ indicates predicting 3D map elements directly. The results in \colorbox{mygray}{gray} are taken from corresponding papers.}
    \begin{tabular}{c|c|cccc|cccc}
    \toprule
    \makebox[2cm][c]{\multirow{2}{*}{Method}} & 
    \makebox[1cm][c]{\multirow{2}{*}{Dim}} & 
    \makebox[1cm][c]{AP$_{div}$} & \makebox[1cm][c]{AP$_{ped}$} & \makebox[1cm][c]{AP$_{bou}$} & \makebox[1cm][c]{mAP$_1$} & \makebox[1cm][c]{AP$_{div}$} & \makebox[1cm][c]{AP$_{ped}$} & \makebox[1cm][c]{AP$_{bou}$} & \makebox[1cm][c]{mAP$_2$} \\
          &       & \multicolumn{4}{c|}{\{0.2m, 0.5m, 1.0m\}} & \multicolumn{4}{c}{\{0.5m, 1.0m, 1.5m\}} \\
    \midrule
    PivotNet \cite{ding2023pivotnet} & \multirow{4}{*}{2} & \cellcolor[rgb]{ .906,  .902,  .902}47.5 & \cellcolor[rgb]{ .906,  .902,  .902}31.3 & \cellcolor[rgb]{ .906,  .902,  .902}43.4 & \cellcolor[rgb]{ .906,  .902,  .902}40.7 & -  & - & - & -\\
    MapTR \cite{liao2022maptr} &       & 40.5  & 27.9  & 32.9  & 33.7  & \cellcolor[rgb]{ .906,  .902,  .902}58.7 & \cellcolor[rgb]{ .906,  .902,  .902}55.4 & \cellcolor[rgb]{ .906,  .902,  .902}59.1 & \cellcolor[rgb]{ .906,  .902,  .902}57.8 \\
    MapTRv2 \cite{liao2023maptrv2} &       & 53.0    & 34.2  & 38.8  & 42.0    & \cellcolor[rgb]{ .906,  .902,  .902}72.1 & \cellcolor[rgb]{ .906,  .902,  .902}62.9 & \cellcolor[rgb]{ .906,  .902,  .902}67.1 & \cellcolor[rgb]{ .906,  .902,  .902}67.4 \\
    \textbf{MapQR} &       & 56.3  & 36.5  & 42.5  & \textbf{45.1} & 72.3  & 64.3  & 68.1  & \textbf{68.2} \\
    \midrule
    MapTRv2 \cite{liao2023maptrv2} & \multirow{2}{*}{3} & 48.1  & 30.5  & 36.7  & 38.4  & \cellcolor[rgb]{ .906,  .902,  .902}69.1 & \cellcolor[rgb]{ .906,  .902,  .902}59.8 & \cellcolor[rgb]{ .906,  .902,  .902}65.3 & \cellcolor[rgb]{ .906,  .902,  .902}64.7 \\
    \textbf{MapQR} &       & 49.4  & 30.6  & 38.9  & \textbf{39.6} & 71.2  & 60.1  & 66.2  & \textbf{65.9} \\
    \bottomrule
    \end{tabular}
  \label{tab:av2}
\end{table}

{\bf Training and Inference Details.}
We follow most settings as in the MapTR series~\cite{liao2022maptr,liao2023maptrv2}, and the modifications will be emphasized. The BEV feature is set to be $200\times100$ to perceive [$-30m, 30m$] for rear to front, and [$-15m, 15m$] for left to right. We use $100$ instance queries to detect map element instances, and each instance is modelled by $20$ sequential points. (We set $N = 100$ and $n = 20$.)

Our model is trained on $8$ NVIDIA 4090 GPUs with a batch size of $8\times4$. The learning rate is set to $6 \times 10^{-4}$. We adopt ResNet50~\cite{he2016deep} as the main backbone to compare with other methods. We also provide results with stronger backbones, including SwinB and SwinT~\cite{liu2021swin}.

\subsection{Comparisons with State-of-the-art Methods}

\begin{figure*}[t] 
\centering 
\includegraphics[width=\linewidth]{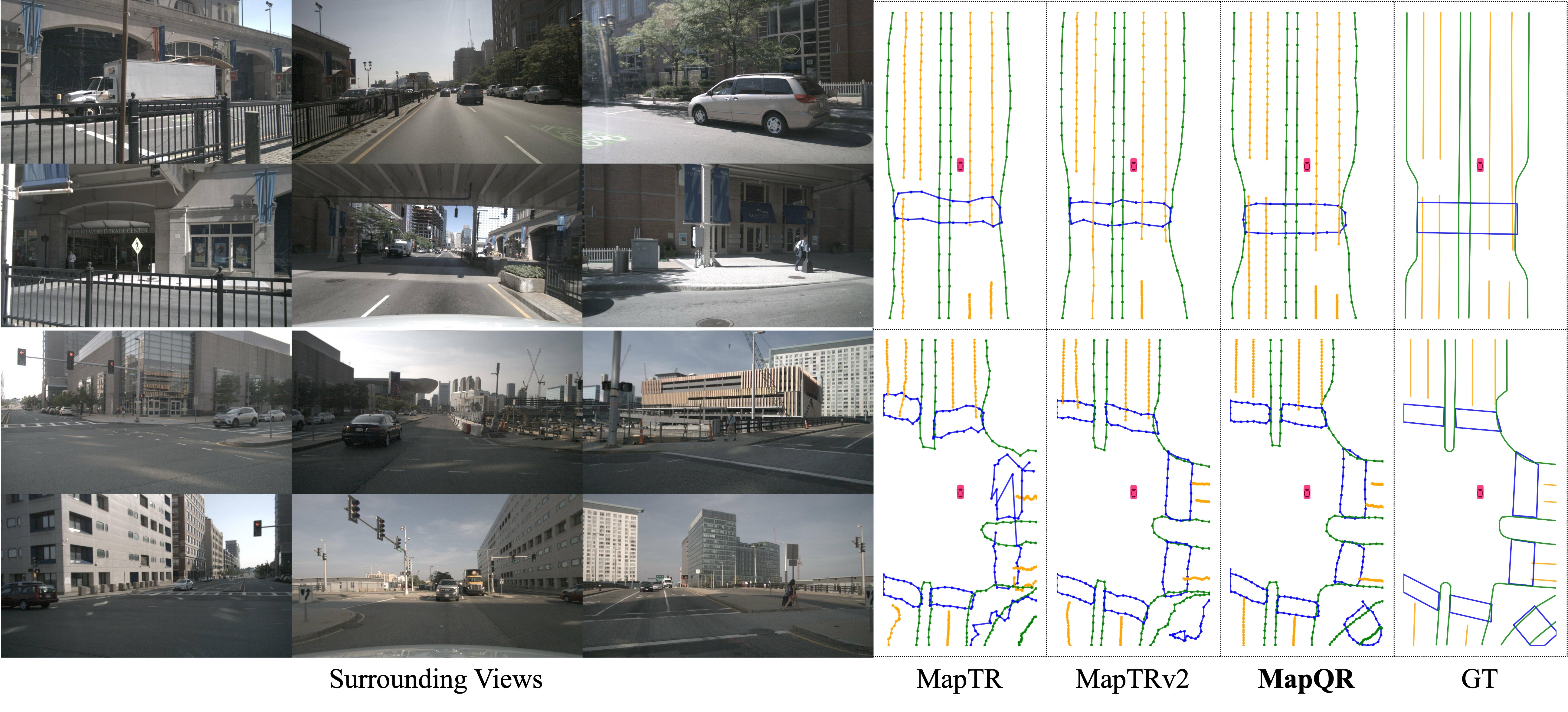}
\caption{Comparison with SOTAs on qualitative visualization. The images are taken from the nuScenes dataset. The \textcolor{orange}{orange}, \textcolor{blue}{blue} and \textcolor[rgb]{0.0235,0.5137,0.0235}{green} colors represent \textcolor{orange}{lane divider}, \textcolor{blue}{pedestrian crossing} and \textcolor[rgb]{0.0235,0.5137,0.0235}{road boundary}, respectively. The proposed method obtains more accurate maps. 
The backbone R50 and $110$ epochs are used in all the methods.
}
\label{fig:rslt-vis}
\end{figure*}

{\bf Results on nuScenes.} 
Following the previous experiment settings, 2D vectorized map elements are predicted by different methods. The experimental results are listed in \cref{tab:nuscenes} and \cref{tab:nuscenes_2}.  
It can be seen that the proposed MapQR outperforms all other SOTA methods by a large margin for both mAP$_1$ and mAP$_2$, under the same settings (i.e., backbone and training epoch).
Compared with MapTR and MapTRv2, BeMapNet~\cite{qiao2023end} and PivotNet~\cite{ding2023pivotnet} get better results under smaller thresholds (i.e., mAP$_1$) thanks to their elaborate modeling. Still, the proposed method without elaborate modeling for map elements can outperform all of them. It implies that our instance query and improved encoder benefit for more accurate prediction. 
The inference speed of our method is about $18$ FPS, which meets the efficiency requirements of many scenarios.

Some qualitative results are shown in \cref{fig:rslt-vis}. In the first row, the comparison methods have acceptable results for forward sensing, while only the proposed method obtains satisfactory results for surround sensing. The second row demonstrates a complex intersection scenario. Our method obtains more satisfactory results than comparison methods. More results can be found in the supplementary materials.

{\bf Results on Argoverse 2.} 
In \cref{tab:av2}, we provide the experimental results on Argoverse 2. These testings are all trained $6$ epochs with ResNet50 as the backbone. Since Argoverse 2 provides 3D vectorized map elements as ground truth, 3D map elements can be predicted directly ($dim=3$) as in MapTRv2. For both 2D and 3D prediction, the proposed MapQR achieves the best performance, especially for stricter thresholds (mAP$_1$). More detailed experimental results are included in the supplementary material. 

\begin{table}[t]
  \centering
  \caption{Integrate the SGQ decoder into other DETR-like map construction models. FPS is tested on the same NVIDIA 4090 machine.}
    \begin{tabular}{c|cc|c}
    \toprule
    \makebox[4cm][c]{Method} & \makebox[2.2cm][c]{mAP$_1$}   & \makebox[2.2cm][c]{mAP$_2$} & \makebox[2cm][c]{FPS} \\
    \midrule
    MapTR & 27.3  & 50.3  & 24.2 \\
    MapTR+SGQ & \textbf{34.4\upperf{7.1}} & \textbf{58.0\upperf{7.7}} & 22.4 \\
    \midrule
    MapTRv2 & 37.2  & 61.5  & 19.6 \\
    MapTRv2+SGQ & \textbf{40.1\upperf{2.9}} & \textbf{63.4\upperf{1.9}} & 19.5 \\
    \midrule
    StreamMapNet & 36.7  & 61.1  & 19.9 \\
    StreamMapNet+SGQ & \textbf{38.4\upperf{1.7}} & \textbf{62.1\upperf{1.0}} & 18.2 \\
    \bottomrule
    \end{tabular}
  \label{tab:ab-decoder}
\end{table}

\begin{table}[t]
  \centering
  \caption{Ablations about scatter-and-gather query and positional embedding. Scatter \& Gather denotes using scatter-and-gather queries. Pos. Embed. denotes positional embed. from reference points. The default settings for our method are highlighted.}
    \begin{tabular}{cc|cc}
    \toprule
    \makebox[3cm][c]{Scatter \& Gather} & \makebox[3cm][c]{Pos. Embed.} & \makebox[2cm][c]{mAP$_1$}  & \makebox[2cm][c]{mAP$_2$} \\
    \midrule
    -     & -     & 28.7  & 51.3 \\
    -     & \checkmark     & 29.9  & 52.6 \\
    \checkmark     & -     & 32.8  & 56.2 \\
    \rowcolor{mypink} \checkmark     & \checkmark     & \textbf{34.4}  & \textbf{58.0} \\
    \bottomrule
    \end{tabular}
  \label{tab:pos_embedding}
\end{table}

\subsection{Ablation Study}
\label{sec:ablation}
We present several ablation experiments for further analysis of our proposed method. All of these experiments are tested on nuScenes dataset with $24$ epochs.

{\bf Decoder Components}. 
To validate the effectiveness of the proposed SGQ decoder, it is integrated into MapTR~\cite{liao2022maptr}, MapTRv2~\cite{liao2023maptrv2} and StreamMapNet~\cite{yuan2024streammapnet_wacv}, and all other settings except for the decoder are kept unchanged. The corresponding methods are denoted as MapTR+SGQ, MapTRv2+SGQ and StreamMapNet+SGQ, respectively. Their performance and relative improvements are also listed in \cref{tab:ab-decoder}.
For MapTR+SGQ, when trained with ResNet50 for $24$ epochs, our decoder helps to improve over $7.0\%$ mAP for both thresholds. Under other settings, the SGQ decoder can also enhance MapTR significantly. While MapTRv2~\cite{liao2023maptrv2} uses decoupled self-attention in the decoder and some auxiliary supervision to improve results, simply replacing their decoder with ours can still step further, especially for the evaluation under stricter thresholds.
For inference speed, introducing our decoder slows down only a little compared with original MapTR or MapTRv2.  
Similar improvements are achieved for StreamMapNet+SGQ.

To further demonstrate the effectiveness of two components in our decoder, we provide an ablation study of scatter-and-gather query and positional embedding in \cref{tab:pos_embedding}. The first row corresponds to the original MapTR, which uses point queries and learnable positional embedding. In the second row, changing the learnable positional embedding to $\{P_{i,j}\}$ described by \cref{eq:pe} can improve the performance. 
In the third row, using the proposed scatter-and-gather query instead of the point query improves the performance significantly.
Adding positional embedding $\{P_{i,j}\}$ of reference points can further improve the performance.

%%%%%%%%%%%%%%%%%
\begin{table}[t]
  \centering
  \caption{Ablations about the number of instance queries ($N_q$). Mem. measures the memory usage for each GPU when training the networks. Blank denotes the values are not available due to the memory limit of NVIDIA 4090 (i.e., $24$ GB). The default settings for our method are highlighted.}
    \begin{tabular}{c|c|cccc}
    \toprule
    \makebox[3cm][c]{Method} & \makebox[1.3cm][c]{$N_q$} & \makebox[1.8cm][c]{mAP$_1$} & \makebox[1.8cm][c]{mAP$_2$}   & \makebox[1.8cm][c]{Mem. (GB)} & \makebox[1.8cm][c]{FPS} \\
    \midrule
    \multirow{4}{*}{MapTR} & 50    & 27.3  & 50.3  & 13.46 & 24.2 \\
          & 75    & 28.2  & 51.1  & 16.31 & 24.2 \\
          & 100   & 28.7  & 51.3  & 20.44 & 22.8 \\
          & 125   & -     & -     & -     & - \\
    \midrule
    \multirow{4}{*}{MapTR+SGQ} & 50    & 32.1  & 55.9  & 11.27  & 22.2 \\
          & 75    & 32.9  & 56.8  & 11.30 & 22.0 \\
          &\cellcolor{mypink}$100$   &\cellcolor{mypink}${\bf 34.4}$  &\cellcolor{mypink}${\bf 58.0}$   &\cellcolor{mypink}$11.51$  &\cellcolor{mypink}$22.4$ \\
          & 125   & 33.8  & 57.4  & 11.59 & 22.8 \\
    \bottomrule
    \end{tabular}
  \label{tab:query_num}
\end{table}

%%%%%%%%%%%%%%%%%

%%%%%%%%%%%%%%%%%
\begin{table}[t]
  \centering
  \caption{Ablations about BEV encoder. BEVPool+ denotes using auxiliary depth supervision. GKT-h is our improved encoder.}
    \begin{tabular}{c|cccc|cccc|c}
    \toprule
    \makebox[2.5cm][c]{\multirow{2}{*}{BEV Encoder}} & \makebox[1cm][c]{AP$_{div}$} & \makebox[1cm][c]{AP$_{ped}$} & \makebox[1cm][c]{AP$_{bou}$} & \makebox[1cm][c]{mAP$_1$}   & \makebox[1cm][c]{AP$_{div}$} & \makebox[1cm][c]{AP$_{ped}$} & \makebox[1cm][c]{AP$_{bou}$} & \makebox[1cm][c]{mAP$_2$} & \makebox[0.7cm][c]{\multirow{2}{*}{FPS}} \\
          & \multicolumn{4}{c|}{\{0.2m, 0.5m, 1.0m\}} & \multicolumn{4}{c|}{\{0.5m, 1.0m, 1.5m\}} &  \\
    \midrule
    BEVFormer~\cite{li2022bevformer} & 39.3  & 29.6  & 33.8  & 34.2  & 60.1  & 54.0    & 59.7  & 57.9  & 22.4 \\
    BEVPool~\cite{huang2022bevpoolv2} & 33.6  & 28.7  & 31.2  & 31.2  & 55.5  & 52.6  & 57.0    & 55.0    & 21.8 \\
    BEVPool+~\cite{huang2022bevpoolv2} & 36.8  & 31.7  & 35.2  & 34.6  & 58.5  & 56.0    & 61.5  & 58.7  & 21.8 \\
    GKT~\cite{chen2022efficient}   & 38.4  & 31.6  & 33.2  & 34.4  & 59.1  & 56.3  & 58.7  & 58.0    & 22.4 \\
    \textbf{GKT-h} & 40.8  & 31.4  & 34.4  & \textbf{35.5} & 61.7  & 56.6  & 60.5  & \textbf{59.6} & 22.3 \\
    \bottomrule
    \end{tabular}
  \label{tab:bev_encoder}
\end{table}

%%%%%%%%%%%%%%%%%

{\bf Instance Query Number}. We provide an ablation study about the instance query number $N$ in \cref{tab:query_num}. The testings are based on our proposed decoder implemented in MapTR. For a fair comparison, we also provide the same ablation results for the original MapTR. The computation complexity of self-attention in its decoder is $O(N^2)$. In our decoder, $N$ is the number of instance queries $N_q$. By contrast, in MapTR $N$ is the number of point queries, which is $N_q \times N_p$, and $N_p$ is the number of points to represent each instance. Therefore, as instance query number $N_q$ increases, the memory usage of MapTR increases dramatically. Although MapTRv2 has realized this problem and proposes a decoupled self-attention, it still consumes nearly all of the GPU memory ($24$ GB) even with $N_q = 50$. In our decoder, lots of memory is saved since self-attention is only among instance queries. The memory usage has negligible increases as $N_q$ increases.

The performance of MapTR increases a little as $N_q$ increases, while our decoder obtains obviously better results for a proper larger $N_q$. 
When the same number of instance queries are used, our decoder outperforms MapTR. Since our method support a larger number of instance queries, we recommend using more instance queries than MapTR to obtain better results.

{\bf BEV Encoder}. This ablation experiment is conducted on MapTR+SGQ by changing different BEV encoders. 
We test commonly used BEV encoders in \cref{tab:bev_encoder}. 
When one encoder layer is used in these BEV encoders, BEVFormer, BEVPool, BEVPool+, and GKT achieve comparable results. Our improved GKT with more flexible height sample locations, denoted as GKT-h, outperforms all other encoders obviously. 

{\bf Performance Gain of Decoder \& Encoder}. From \cref{tab:pos_embedding}, the SGQ decoder plays a more significant role in performance gain, contributing an increase of $6.7$ mAP$_2$ points. From \cref{tab:bev_encoder}, the GKT-h encoder, which contributes a relative smaller improvement of $1.6$ points, is a simple and effective enhancement.
\section{Conclusion}
\label{sec:conclusion}
In this paper, we explore the query mechanism for better performance in online map construction task.
Inspired by the frontier research in DETR-like architecture~\cite{meng2021conditional,liu2022dab}, we design a novel scatter-and-gather query for the decoder. Therefore, in cross-attention, each point query for the same instance shares the same content information with different positional information, embedded from different reference points. 
We demonstrate the performance of SOTA methods can be further improved by combining them with our decoder. With our improvements in a BEV encoder, our new framework MapQR also achieves the best results in both nuScenes and Argoverse 2.

\bibliographystyle{splncs04}
\bibliography{main}

\newpage
\appendix
% ---------------------------------------------------------------
%\newpage
\definecolor{mypink}{rgb}{.99,.91,.95}
\section*{Appendix}
\section{Motivation}

\begin{figure}[h]
\centering 
\setlength{\abovecaptionskip}{0.cm}
\includegraphics[width=0.9\linewidth]{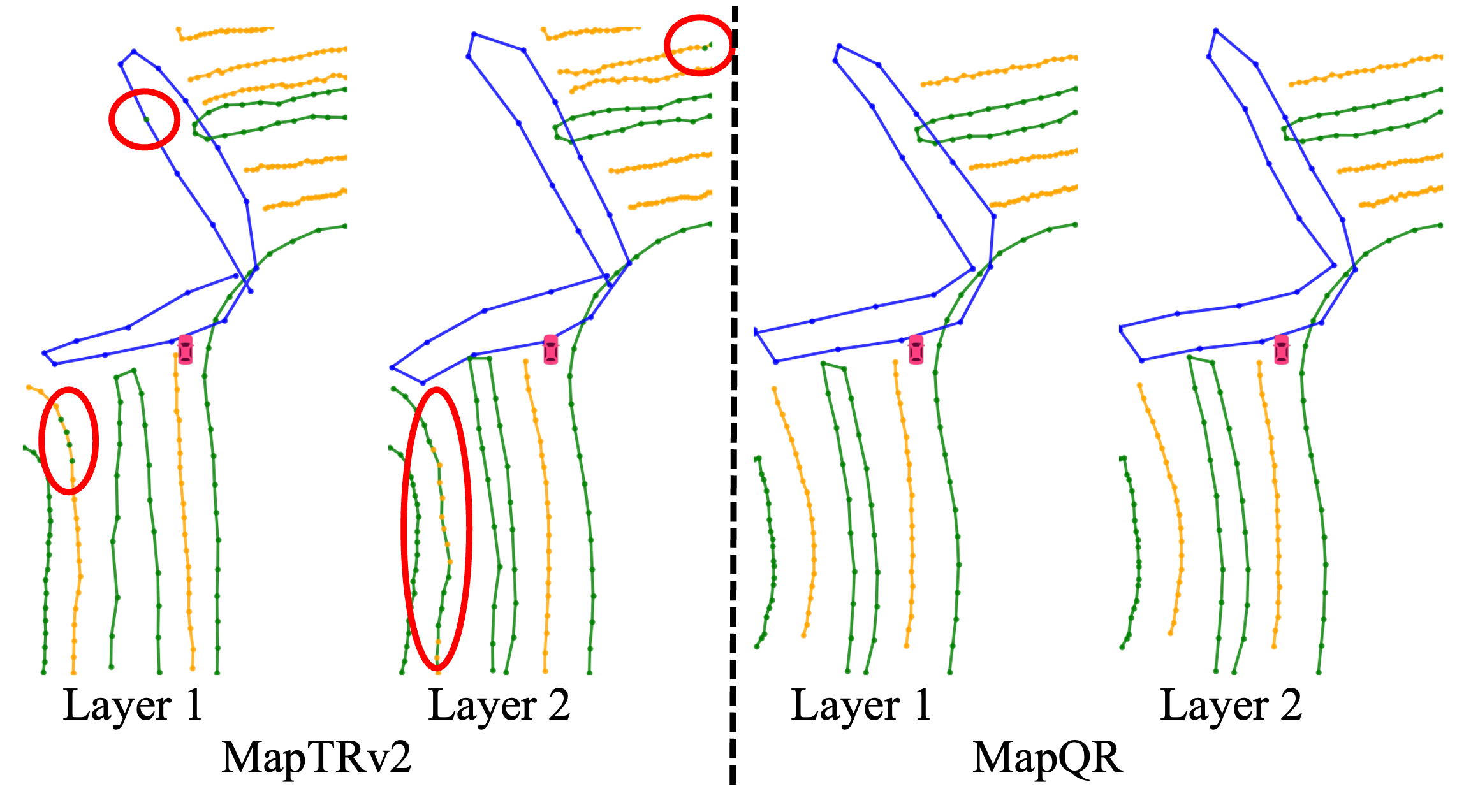}
\vspace{-0em}
\caption{
\footnotesize Output of the first two decoder layers. The point and instance are colored by their predicted labels.  {\bf Left}: In MapTRv2, instance labels are predicted from the average of point queries. 
Point queries within the same instance may have different labels, resulting content conflict. See red circles. 
{\bf Right}: Since the proposed SGQ ensures shared content within the same instance, MapQR can completely avoid such conflict.}
\label{fig:label}
\vspace{-1.em}
\end{figure}

% \subsection{Content Conflict in Point-Based-Query} 
Since modelled as a point set prediction problem, it is straightforward to use point queries in a transformer architecture. 
Like in MapTR~\cite{liao2022maptr}, each point query will predict a position and they will be grouped together in a preset order to form final predicted map elements. 
Based on Deformable DETR~\cite{zhu2021deformable}, each point query is probing BEV features around its reference point. 
Although features will be exchanged and enhanced in self-attention module, it is still hard for a point query to contain all content information of the whole map element. This is verified in \cref{fig:label}, in which different type of map elements are drawn in different colors. 
Besides map element type, here we also draw point type which is predicted from each point query directly. 
We can find some points (emphasized in red circles) are predicted different labels although they are within the same map element.
This reveals that if point queries are directly used to construct the final map element instance, there will be content conflicts within the same instance.

The motivation of our method is overcoming the limitation of point query in point set prediction, which is popular in mainstream end-to-end online map construction. 

\section{More Experimental Results}
\subsection{Detailed Experimental Results} 
In  \cref{tab:query_num-supp}, more detailed experimental results are provided, including the average precision (AP) for lane divider, pedestrian crossing, and road boundary respectively. These results are consistent with Tab.~2 to Tab.~5 in the main manuscript. 

\subsection{Convergence Curve}
\begin{figure}[t] 
	\centering 
	\includegraphics[width=8cm]{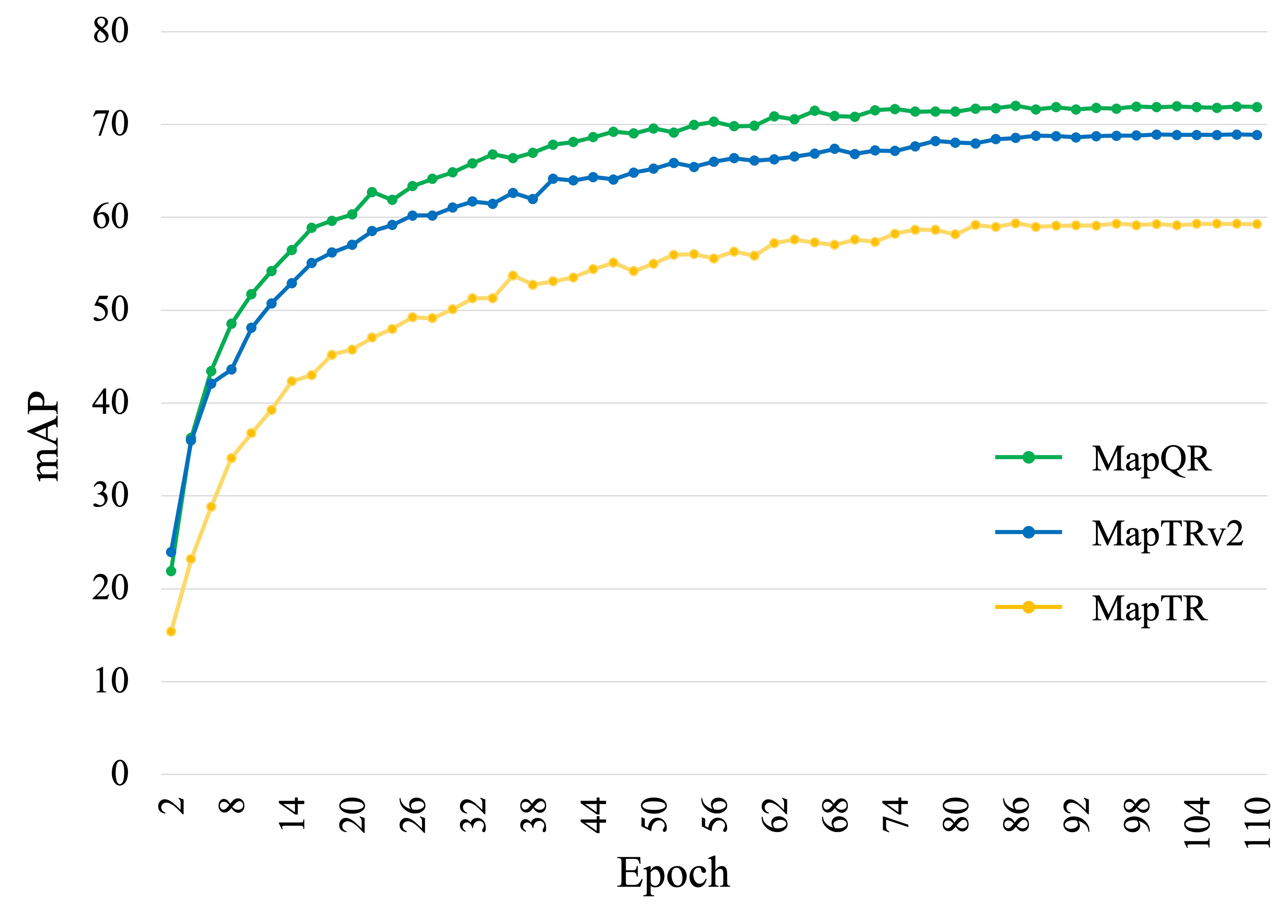}
	\caption{Convergence curves of MapQR, MapTR, and MapTRv2. These models are all trained 110 epochs, with ResNet50 as backbone. }
	\label{fig:curve}
\end{figure}

We provide the training convergence curves in \cref{fig:curve}. Like the MapTR series, the proposed MapQR has converged when trained $110$ epochs, and achieves obviously better performance than MapTR~\cite{liao2022maptr} and MapTRv2~\cite{liao2023maptrv2}.

\subsection{Ablation about Positional Embedding for Instance Queries}
\begin{figure}[t] 
	\centering 
	\includegraphics[width=8cm]{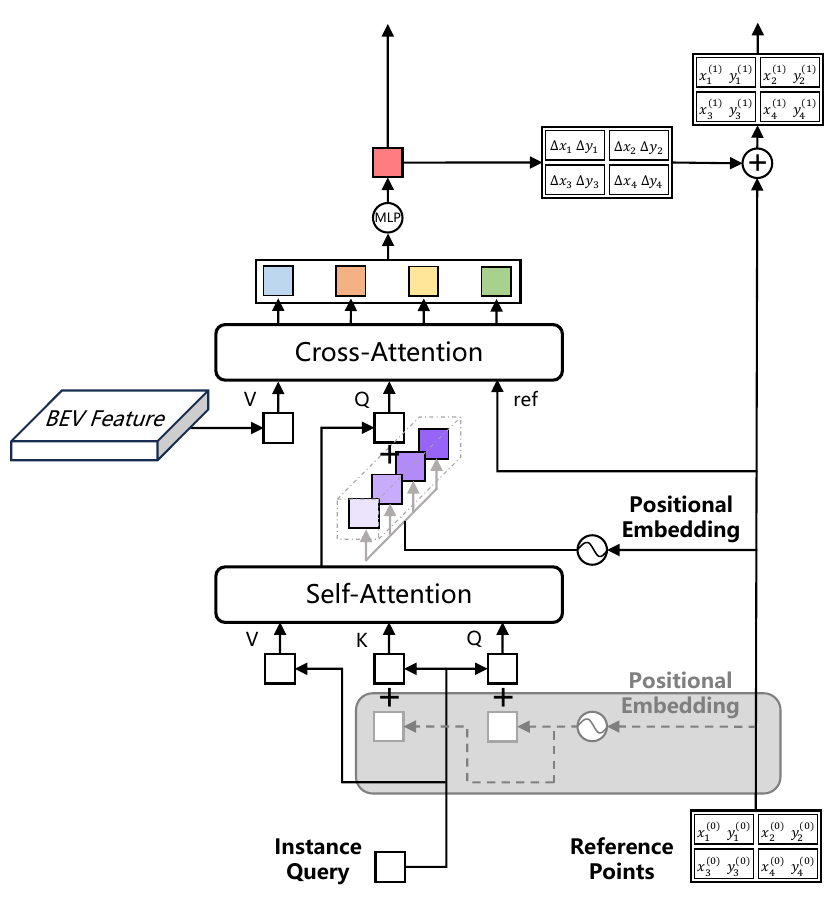}
	\caption{Structure of the proposed MapQR with positional embedding for instance queries, which is highlighted in the gray block. We do not use this positional embedding for the default setting.}
	\label{fig:instance_embed}
\end{figure}

In \cref{tab:inst_pe}, we provide an ablation study about the positional embedding for instance queries. This positional embedding is added to each instance query before self-attention, as illustrated in \cref{fig:instance_embed}. We have tested some common choices for generating positional embedding (PE), including no PE, bounding box like in DAB-DETR~\cite{liu2022dab}, center point like in Conditional DETR~\cite{meng2021conditional}, and learnable PE like in original MapTR~\cite{liao2022maptr}. The best results are obtained when no PE is used. Thus, we do not include any PE for instant queries in our model.

\subsection{Ablation about Number of BEV Encoder Layer}
As indicated in PivotNet~\cite{ding2023pivotnet}, more encoder layers can also bring benefits. Thus we also provide an ablation study about the number of BEV encoder layer on MapQR in \cref{tab:bev_layer}. As the number of layer increasing, the performance is also improved while with some cost of efficiency.
To achieve an accuracy-and-efficiency trade-off, we set the layer number as $3$ for the default setting.

\subsection{Qualitative Results}
\begin{figure*}[t] 
\centering 
\includegraphics[width=0.95\linewidth]{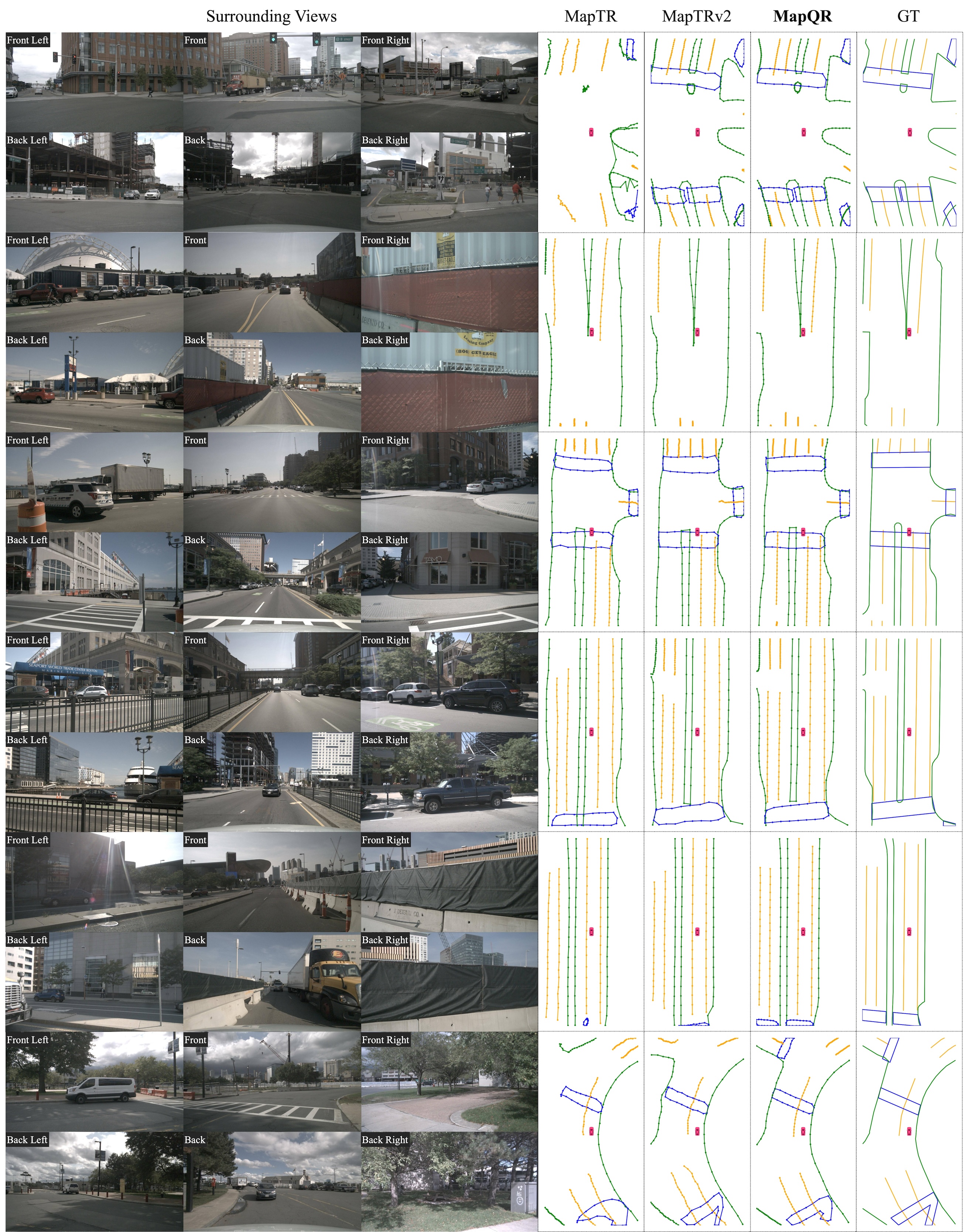}
\caption{The \textcolor{orange}{orange}, \textcolor{blue}{blue} and \textcolor[rgb]{0.0235,0.5137,0.0235}{green} colors represent \textcolor{orange}{lane divider}, \textcolor{blue}{pedestrian crossing} and \textcolor[rgb]{0.0235,0.5137,0.0235}{road boundary}, respectively.
}
\label{fig:rslt-vis1}
\end{figure*}

\begin{figure*}[t] 
\centering 
\includegraphics[width=0.95\linewidth]{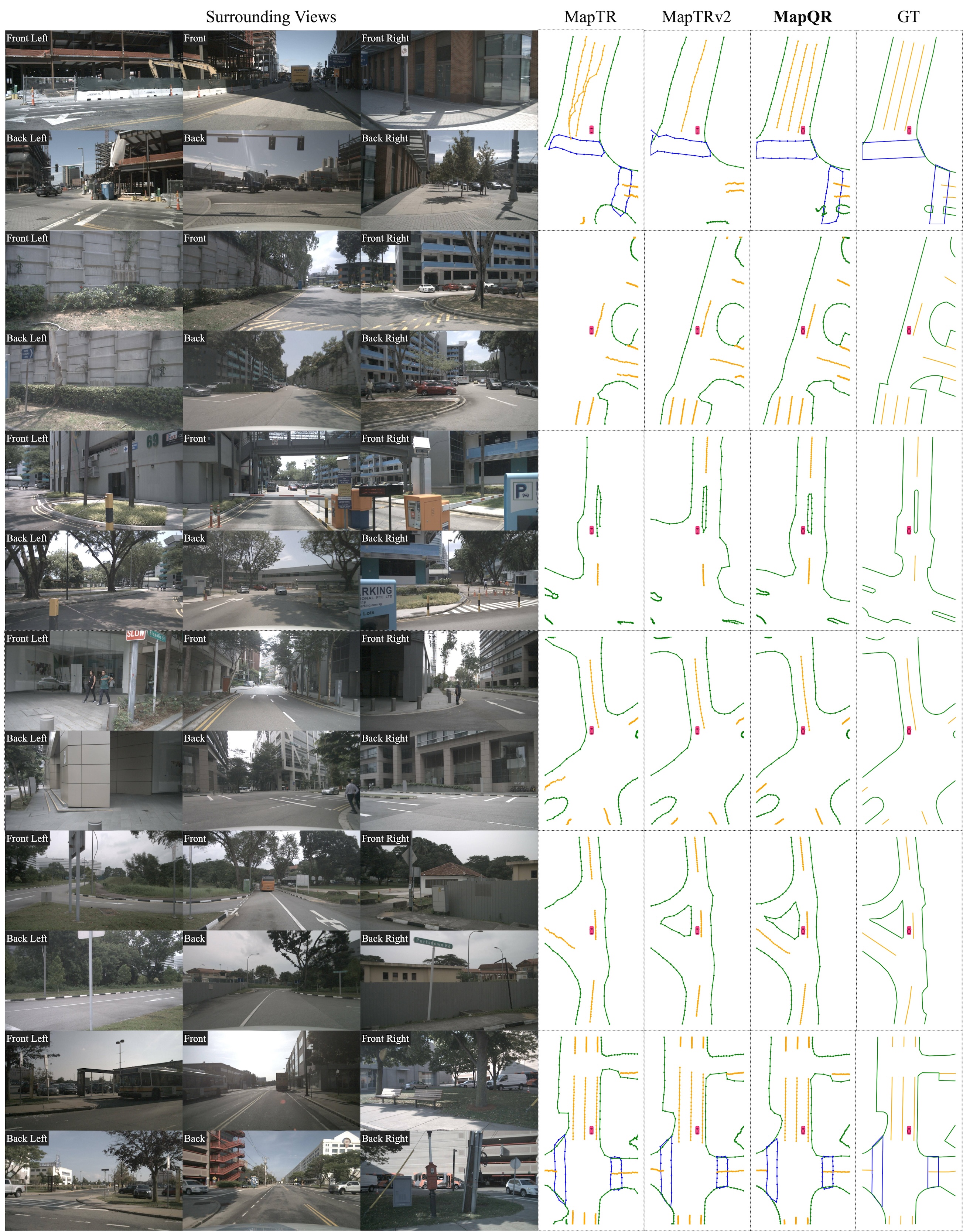}
\caption{The \textcolor{orange}{orange}, \textcolor{blue}{blue} and \textcolor[rgb]{0.0235,0.5137,0.0235}{green} colors represent \textcolor{orange}{lane divider}, \textcolor{blue}{pedestrian crossing} and \textcolor[rgb]{0.0235,0.5137,0.0235}{road boundary}, respectively.
}
\label{fig:rslt-vis2}
\end{figure*}

\begin{figure*}[t] 
\centering 
\includegraphics[width=0.95\linewidth]{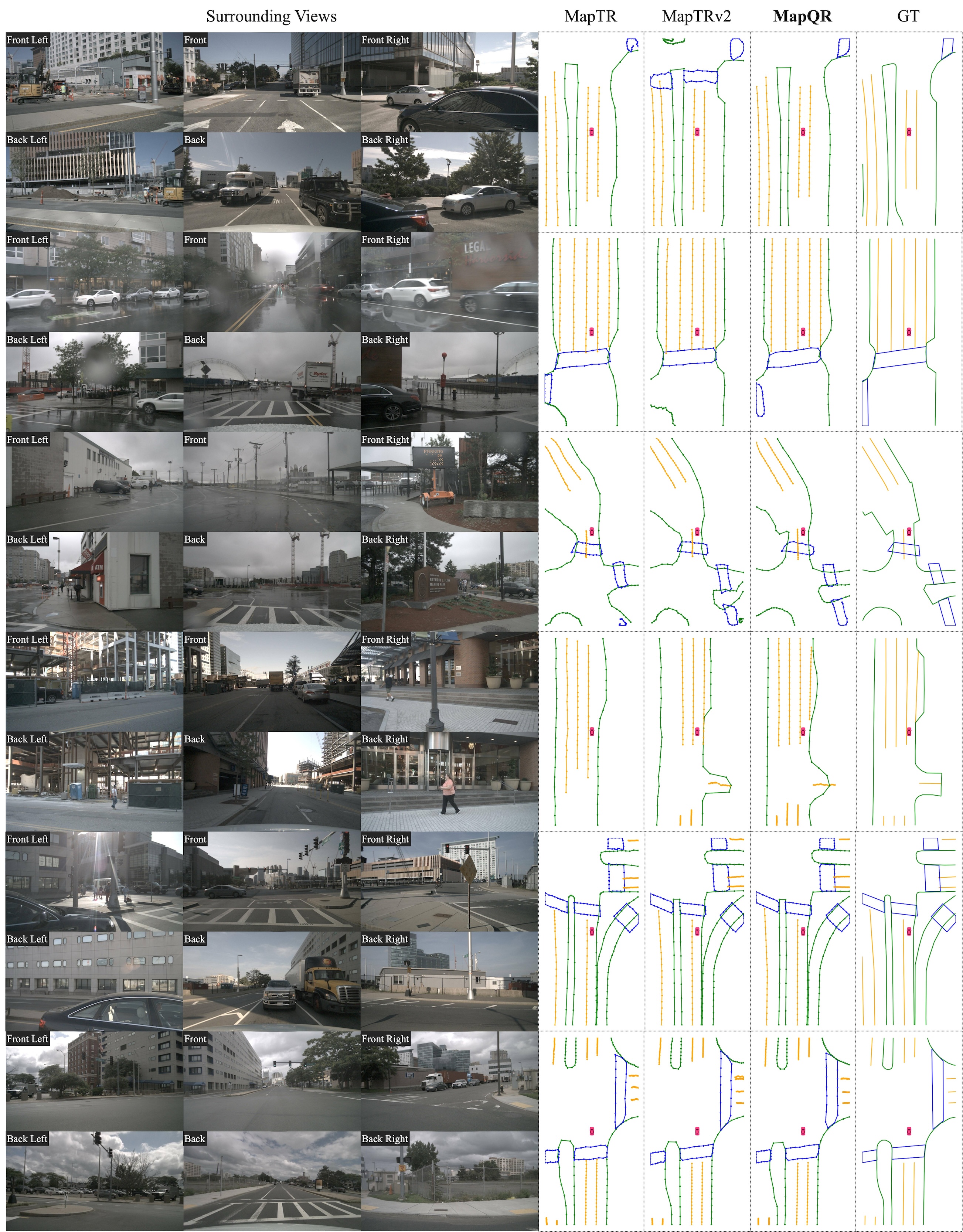}
\caption{The \textcolor{orange}{orange}, \textcolor{blue}{blue} and \textcolor[rgb]{0.0235,0.5137,0.0235}{green} colors represent \textcolor{orange}{lane divider}, \textcolor{blue}{pedestrian crossing} and \textcolor[rgb]{0.0235,0.5137,0.0235}{road boundary}, respectively.
}
\label{fig:rslt-vis3}
\end{figure*}

We provide more qualitative results on nuScenes~\cite{caesar2020nuscenes} in \cref{fig:rslt-vis1}, \cref{fig:rslt-vis2}, and \cref{fig:rslt-vis3}. For each example, the $3$ images in the first row are taken by the cameras of front-left, front, and front-right; the $3$ images in the second row are taken by the cameras of back-left, back, and back-right. For simple scenarios, all these methods can predict sufficiently accurate map elements. While for complex scenarios, MapQR predicts overall more accurate map elements with less false negatives and false positives.

\begin{table*}[t]
	\centering
	\caption{Ablations about the number of instance queries ($N_q$). Mem. measures the memory usage for each GPU when training the networks. Blank denotes the values are not available due to the memory limit of NVIDIA 4090 (i.e., $24$ GB). The default settings for our method are highlighted.} 
	\begin{tabular}{c|c|cccc|cccc|c}
        \toprule
		\makebox[1.3cm][c]{\multirow{2}{*}{Methods}} & \makebox[0.7cm][c]{\multirow{2}{*}{$N_q$}} & \makebox[1cm][c]{AP$_{{div}}$} & \makebox[1cm][c]{AP$_{{ped}}$} & \makebox[1cm][c]{AP$_{{bou}}$} & \makebox[1cm][c]{mAP$_1$}   & \makebox[1cm][c]{AP$_{{div}}$} & \makebox[1cm][c]{AP$_{{ped}}$} & \makebox[1cm][c]{AP$_{{bou}}$} & \makebox[1cm][c]{mAP$_2$} & \makebox[1cm][c]{Mem.} \\
		&  & \multicolumn{4}{c|}{\{0.2m, 0.5m, 1.0m\}} & \multicolumn{4}{c|}{\{0.5m,  1.0m, 1.5m\}} & (GB)\\
		\midrule
		\multirow{4}{*}{{\begin{tabular}{c} MapTR \\ \cite{liao2022maptr}
		\end{tabular}}} & $50$ & $30.7$ & $23.2$ & $28.2$ & $27.3$ & $51.5$ & $46.3$ & $53.1$ & $50.3$ & $13.46$ \\
		& $75$ & $31.2$ & $23.2$ & $30.1$ & $28.2$ & $51.8$ & $46.3$ & $55.3$ & $51.1$ & $16.31$ \\
		& $100$ & $31.2$ & $24.3$ & $30.8$ & $28.7$ & $51.0$ & $47.3$ & $55.7$ & $51.3$ & $20.44$ \\
		& $125$   & -  & - & -   & - & -  & - & -   & -     & - \\
		\midrule
		\multirow{4}{*}{{\begin{tabular}{c} MapTR \\ {\bf +SGQ}
		\end{tabular}}} & $50$ & $35.3$ & $28.6$ & $32.4$ & $32.1$ & $57.3$ & $52.4$ & $57.9$ & $55.9$  & $11.27$ \\
		& $75$ & $36.8$ & $28.9$ & $32.9$ & $32.9$ & $58.4$ & $53.7$ & $58.4$ & $56.8$  & $11.30$ \\
		&\cellcolor{mypink}$100$   &\cellcolor{mypink}${ 38.4}$  &\cellcolor{mypink}${ 31.6}$ &\cellcolor{mypink}${ 33.2}$  &\cellcolor{mypink}${\bf 34.4}$ &\cellcolor{mypink}${ 59.1}$  &\cellcolor{mypink}${ 56.3}$ &\cellcolor{mypink}${ 58.7}$  &\cellcolor{mypink}${\bf 58.0}$   &\cellcolor{mypink}$11.51$ \\
		& $125$ & $38.9$ & $29.2$ & $33.2$ & $33.8$ & $59.8$ & $53.4$ & $59.0$ & $57.4$  & $11.59$ \\
		\bottomrule
	\end{tabular}
	\label{tab:query_num-supp}
\end{table*}

\begin{table*}[t]
	\centering
	\caption{Ablations about positional embedding for instance queries. These experiments are tested on MapTR+SGQ. ``Without'' denotes not using any positional embedding, which is the default setting for our method. ``Bounding Box'' denotes embedding from the bounding boxes (i.e., $x, y, w, h$) of corresponding map instances. ``Center Point'' denotes embedding from only the center points (i.e., $x, y$) of corresponding bounding boxes. ``Learnable'' denotes using random learnable embedding.
	}
	\begin{tabular}{c|cccc|cccc|c}
		\toprule
		\makebox[2.2cm][c]{\multirow{2}{*}{Embedding}} & \makebox[1cm][c]{AP$_{{div}}$} & \makebox[1cm][c]{AP$_{{ped}}$} & \makebox[1cm][c]{AP$_{{bou}}$} & \makebox[1cm][c]{mAP$_1$}   & \makebox[1cm][c]{AP$_{{div}}$} & \makebox[1cm][c]{AP$_{{ped}}$} & \makebox[1cm][c]{AP$_{{bou}}$} & \makebox[1cm][c]{mAP$_2$}  & \makebox[1cm][c]{\multirow{2}{*}{FPS}} \\
		&  \multicolumn{4}{c|}{\{0.2m, 0.5m, 1.0m\}} & \multicolumn{4}{c|}{\{0.5m, 1.0m, 1.5m\}} \\
		\midrule
        \rowcolor{mypink} Without & $38.4$ & $31.6$ & $33.2$ & $\bf{34.4}$ & $59.1$ & $56.3$ & $58.7$ & $\bf{58.0}$ & $22.4$  \\
		Bounding Box & $37.5$ & $30.8$ & $31.4$ & $33.2$ & $58.8$ & $54.5$ & $56.8$ & $56.7$ & $20.5$\\
		Center Point & $37.4$ & $28.6$ & $31.2$ & $32.4$ & $58.8$ & $53.2$ & $56.6$ & $56.2$ & $21.2$\\
		Learnable &  ${38.4}$  &${30.4}$ &${33.5}$  &${34.1}$ &${58.8}$  &${55.5}$ &${59.4}$  &${57.9}$  & $22.3$   \\
		\bottomrule
	\end{tabular}
	\label{tab:inst_pe}%
\end{table*}%

\begin{table*}[t]
	\centering
	\caption{Ablations about the number of BEV encoder layers (\#layer). The default settings for our method are highlighted.
	}
	\begin{tabular}{c|cccc|cccc|c}
		\toprule
		\makebox[1.3cm][c]{\multirow{2}{*}{\#layer}} & \makebox[1.1cm][c]{AP$_{div}$} & \makebox[1.1cm][c]{AP$_{ped}$} & \makebox[1.1cm][c]{AP$_{bou}$} & \makebox[1.1cm][c]{mAP$_1$}   & \makebox[1.1cm][c]{AP$_{div}$} & \makebox[1.1cm][c]{AP$_{ped}$} & \makebox[1.1cm][c]{AP$_{bou}$} & \makebox[1.1cm][c]{mAP$_2$}  & \makebox[1cm][c]{\multirow{2}{*}{FPS}} \\
		&  \multicolumn{4}{c|}{\{0.2m, 0.5m, 1.0m\}} & \multicolumn{4}{c|}{\{0.5m, 1.0m, 1.5m\}} \\
		\midrule
		1 & $46.4$ & $36.1$ & $39.7$ & $40.8$ & $64.8$ & $60.9$ & $65.2$ & $63.6$ & $21.3$  \\
		2 & $49.3$ & $38.0$ & $40.4$ & $42.6$ & $66.9$ & $62.5$ & $66.4$ & $65.3$ & $19.8$\\
        \rowcolor{mypink} 3 & $49.9$ & $38.6$ & $41.5$ & $43.3$ & $68.0$ & $63.4$ & $67.7$ & $66.4$ & $17.9$\\
        4 & ${50.3}$  &${39.9}$ &${41.0}$  &${43.8}$ &${69.2}$  &${65.6}$ &${67.3}$  &${67.4}$ & $16.9$\\
		\bottomrule
	\end{tabular}
	\label{tab:bev_layer}
\end{table*}

\label{sec:supp-exp}

\end{document}